\documentclass[conference]{IEEEtran}

\usepackage{cite}
\usepackage{amsmath,amssymb,amsfonts}
\usepackage{algorithmic}
\usepackage{graphicx}
\usepackage{booktabs}
\usepackage{multirow}
\usepackage{array}
\usepackage{textcomp}
\usepackage{xcolor}
\usepackage{url}
\usepackage{balance}
\usepackage{tikz}
\usepackage{pgfplots}
\usepackage{pgfplotstable}
\usepackage{caption}

\pgfplotsset{compat=1.18}
\usetikzlibrary{positioning,arrows.meta,fit,shapes.geometric,calc}

\makeatletter
\newcommand{\IEEEcopyrightfooter}{%
  \copyright2026 IEEE.  Personal use of this material is permitted.  Permission from IEEE must be obtained for all other uses, in any current or future media, including reprinting/republishing this material for advertising or promotional purposes, creating new collective works, for resale or redistribution to servers or lists, or reuse of any copyrighted component of this work in other works.%
}

\def\ps@IEEEtitlepagestyle{%
  \def\@oddhead{}%
  \def\@evenhead{}%
  \def\@oddfoot{%
    \hbox to \textwidth{%
      \hfil
      \parbox{0.98\textwidth}{%
        \centering\fontsize{8}{9}\selectfont\IEEEcopyrightfooter
      }%
      \hfil
    }%
  }%
  \def\@evenfoot{}%
}
\makeatother

\newcommand{\sys}{\mbox{AgentRiskBOM}}

\def\BibTeX{{\rm B\kern-.05em{\sc i\kern-.025em b}\kern-.08em
    T\kern-.1667em\lower.7ex\hbox{E}\kern-.125emX}}

\begin{document}

\title{AgentRiskBOM: A Risk-Scoping Security Bill of Materials for Agentic AI Systems}

\author{\IEEEauthorblockN{Srimonti Dutta}
\IEEEauthorblockA{\textit{WAI USA Research Labs} \\
\textit{WAI USA}\\
Austin, USA \\
srimonti@womeninai.co}
\and
\IEEEauthorblockN{Akshata Kishore Moharir}
\IEEEauthorblockA{\textit{WAI USA Research Labs} \\
\textit{WAI USA}\\
Seattle, USA \\
akshata@womeninai.co}
}

\maketitle

\begin{abstract}
Agentic AI systems are no longer passive model endpoints: they retrieve private context, invoke tools, write files, call external services, coordinate with other agents, and may act without human approval. Existing bill-of-materials artifacts improve transparency for software dependencies, model metadata, and training provenance, but they leave a distinct agentic transparency gap: \textit{capability opacity}, the absence of a structured account of what a deployed agent can access, remember, change, delegate, and prove after the fact. This paper introduces \sys{}, a security bill of materials for risk-scoping tool-using AI agents. \sys{} is designed as an additive layer over SBOM, AI-BOM, and ML-BOM artifacts: it references those artifacts where they are authoritative, while adding fields for agent-specific runtime authority, including autonomy, tool permissions, memory, credential scope, approval gates, audit signals, inter-agent communication, and external action capability.

We implement \sys{} as a JSON-schema-based artifact with a reproducible corpus, risk-scenario library, rule-based scorer, diff detector, control mapper, and rendered reports. The evaluation uses 13 documented open-source agents spanning coding, retrieval-augmented generation (RAG), and multi-agent archetypes, together with 52 risk scenarios across 14 categories. The schema validates all 13 corpus artifacts. Prior-art coverage analysis gives \sys{} a native-equivalent coverage score of 14 across 16 capability dimensions, compared with 1.0 for SBOM, 1.5 for AI-BOM, and 2.0 for ML-BOM under the same matrix. Across applicable modeled risk categories, \sys{} exposes 100.0\% of risk-category visibility, compared with 10.5\% for SBOM-like views and 20.9\% for AI-BOM-like views. To test agentic authority drift, we inject 33 structured deployment mutations; the diff detector identifies the correct change type for all mutations. A secondary penalty-based scorer yields a Spearman rank correlation of 0.73 with the primary scorer, supporting rank-level consistency while showing that categorical thresholds require human calibration.
The results show that agentic AI security needs a machine-readable authority-and-risk artifact before runtime incidents occur.
\end{abstract}


\section{Introduction}
A bill of materials is valuable in security because it turns an opaque system into a reviewable inventory. For software, that inventory helps expose dependencies, vulnerabilities, licenses, and supply-chain provenance. For agentic AI, the missing inventory is what the system is allowed to do.  In software engineering, SBOMs (Software Bill of Materials) help consumers understand dependencies, vulnerabilities, license exposure, and supply-chain provenance \cite{b1,b2}.  In machine learning, model cards and datasheets improved transparency around intended use, model behavior, and dataset construction \cite{b3,b4}; more recent AI-BOM (AI Bill of Materials) work extends the BOM model toward trained-model provenance, environments, and lifecycle evidence \cite{b5,b6}.  These artifacts are necessary, but they do not answer the most consequential question for an AI agent: \emph{what is the system allowed to do?}

That question is important because the architecture of AI systems has changed.  Tool-using language agents combine reasoning with action, invoke APIs, read and write files, browse the Web, execute code, and coordinate with other agents \cite{b7,b8,b9,b10}.  Benchmarks such as AgentBench treat LLMs as interactive decision makers rather than text-only predictors \cite{b11}.  Indirect prompt injection, tool-integrated attacks, unsafe external actions, and model-supply-chain risks show that the security boundary around an agent is not captured by software dependencies alone \cite{b12,b13,b14,b15}.  A dependency inventory may reveal that an agent imports a browser library, but it will not reveal whether the agent can send an external email without approval, invoke a shell, write to a repository, retain sensitive memory, or delegate remediation to another agent.

This paper argues that agentic AI systems require a new kind of BOM: \sys{}, a risk-scoping artifact that connects static inventory to delegated authority. \sys{} is not a replacement for SBOMs, AI-BOMs, or ML-BOMs (Machine Learning Bill of Materials), it is a complementary security layer for runtime authority, which makes agentic risk visible before deployment and trackable across changes. \sys{} records agent identity, model and prompt metadata, tool descriptors, tool-risk tiers, autonomy level, memory and data sources, approval gates, audit signals, inter-agent communication, credential scope, control mappings, and references to external BOMs.

The central contribution of this work is therefore a practical risk-scoping substrate for agentic systems.  The paper makes four claims.

\textbf{C1: Structural gap.}  Existing BOM classes provide weak or no native representation for the fields needed to review agentic runtime authority.

\textbf{C2: Expressiveness.}  A lightweight schema can represent real open-source agent designs without forcing every field to apply to every archetype.

\textbf{C3: Lifecycle utility.}  Once agentic authority is represented structurally, risk scoring, diff-based deployment review, control mapping, and forensic-readiness assessment become simple, reproducible computations.

\textbf{C4: Reviewability.}  A useful agent-security artifact should support both human review and automated checks.  \sys{} is designed to make agent authority explicit enough for security reviewers, while remaining structured enough for validation, scoring, diffing, and control mapping in deployment workflows.

\section{Related Work and Research Gap}
\subsection{SBOMs and Supply-Chain Transparency}
One of the first empirical studies of SBOM practice finds that practitioners value SBOM transparency but still face tooling, trust, and adoption challenges \cite{b1}.  A broader stakeholder study similarly shows that BOM production, maintenance, and domain-specific use remain challenging even as SBOMs gain attention \cite{b2}. A landscape study of SBOM tools identifies gaps across both generation and consumption support \cite{b16}.  Supply-chain attestation work addresses a related provenance problem: in-toto provides cryptographic guarantees for software supply-chain integrity and allows end users to verify supply-chain steps from project inception to deployment \cite{b17}.  \sys{} is complementary to such attestation mechanisms because it records the declared authority envelope of a deployed agent rather than only the integrity of software build steps. These studies support the premise that BOMs matter, but they also clarify the boundary of SBOMs: the primary unit of representation is software composition, not the operational authority of an autonomous or semi-autonomous agent.

\subsection{AI-BOMs, Model Documentation, and Provenance}
Model cards and datasheets established documentation patterns for trained models and datasets \cite{b3,b4}.  AI-BOM work extends supply-chain inventory toward algorithms, data collection methods, frameworks, licensing, standards, and model environments \cite{b5,b6}.  AIBoMGen goes further by generating signed AI-BOMs during model training, capturing datasets, model metadata, environment details, and integrity evidence \cite{b6}. Recent AI assurance work also moves beyond descriptive documentation toward evidence-bound verification.  The AI Risk Scanning (AIRS) framework aligns assurance fields with adversarial-ML threat modeling and generates structured artifacts for model integrity, packaging and serialization safety, structural adapters, and runtime behaviors \cite{b18}.  However, AIRS is currently scoped to model-level assurances for LLMs, while system-level threats such as application-layer abuses and tool-calling are identified as possible extensions.  \sys{} is therefore complementary: it focuses directly on deployed agent authority, tools, memory, approval boundaries, and auditability. Supply-chain security work for pre-trained models shows why this lineage matters: model registries and artifacts introduce risks that differ from conventional package ecosystems \cite{b12}.

These works are highly relevant to \sys{}, but they operate mostly at the model, data, training, or software-component layer.  By contrast, \sys{} is concerned with deployed agent authority: tools, permissions, memory, approval gates, inter-agent trust, and auditability.  This is why \sys{} contains external BOM references rather than duplicating SBOM or AI-BOM fields.

\subsection{Tool-Using Agents and Agentic Risk}
ReAct demonstrated the value of interleaving reasoning traces with actions over external environments \cite{b7}.  Toolformer showed that language models can learn to call APIs such as calculators, search engines, translation systems, and calendars \cite{b8}.  AutoGen demonstrates that multi-agent conversation can coordinate multiple agents, human input, LLMs, and tools in flexible applications \cite{b9}.  Surveys of LLM-based agents describe the broader shift from single-model inference to systems that sense, decide, and act \cite{b10}.

The same shift expands the attack surface.  Prompt-injection work on real LLM-integrated applications shows practical attacks such as application prompt theft and unauthorized use of application capabilities \cite{b19}.  Indirect prompt injection can compromise applications that retrieve untrusted external content \cite{b13}.

Benchmarks for indirect prompt injection show that models struggle to separate data from instructions \cite{b14}.  In tool-integrated agent workflows, indirect prompt injection can cause harmful actions and data exfiltration \cite{b15}.  These motivate \sys{} fields such as trusted input boundaries, tool descriptors, approval requirements, retrieval logging, and external action capability.

\subsection{Operational Authority in Tool-Using Agents}
The agent-security literature shows that the main risk shift is operational rather than merely architectural. Indirect prompt injection can cause an application to treat retrieved content as instructions, and tool-integrated agents can translate such instructions into external actions \cite{b13,b14,b15}. Multi-agent frameworks further complicate the boundary of responsibility because tasks, context, and tool access may be distributed across several agents \cite{b9,b10}. These observations motivate a representation that records not only the presence of tools, but their authority, data reachability, approval requirements, and audit evidence.

\sys{} is positioned at this operational layer. It does not attempt to replace execution logs, traces, or incident-response tooling. Instead, it records the declared authority envelope of an agent before deployment and across releases. This distinction is important: many dangerous changes in an agent system are visible as configuration or governance changes before they appear as incidents. A schema that exposes autonomy, tool tier, memory persistence, credential scope, external action capability, and logging posture gives reviewers a concrete object to inspect before the agent is connected to production systems.

\begin{table}[t]
\caption{Security-review questions encoded by \sys{}.}
\label{tab:questions}
\centering
\scriptsize
\begin{tabular}{p{0.33\columnwidth}p{0.57\columnwidth}}
\toprule
\textbf{Question} & \textbf{Representative fields} \\
\midrule
What can the agent do without a human? & Autonomy level, approval rules, max tool tier. \\
What external systems can it affect? & Tool endpoints, side effects, network boundary. \\
What sensitive data can it see or remember? & Data class, RAG sources, memory type, retention. \\
Can risky changes be caught before deployment? & Tool diff, prompt hash diff, approval/logging diff. \\
Could an incident be reconstructed? & Prompt/tool/retrieval/approval logs, retention, tamper resistance. \\
Which controls should be required? & Risk drivers, control-mapping rules, provenance fields. \\
\bottomrule
\end{tabular}
\end{table}

\section{\sys{} Design}

\subsection{Design Goals}
\sys{} follows five design principles.  First, it is \emph{additive}: SBOMs, AI-BOMs, and ML-BOMs remain authoritative for dependency and model-provenance fields.  Second, it is \emph{schema-first}: every reviewable property is represented as a typed field.  Third, it is \emph{risk-scoping}: it focuses on authority, exposure, and governance rather than exhaustive runtime traces.  Fourth, it is \emph{diffable}: changes in autonomy, tools, memory, logging, and approval gates should be visible to CI/CD gates.  Fifth, it is \emph{audit-oriented}: the artifact records whether post-incident reconstruction would be possible.

\subsection{Threat Model and Review Questions}
The threat model reflects practical deployment conditions. We assume that the base model, agent framework, and tools may be supplied by different parties; that retrieved content may be untrusted; that tool descriptions may be stale or maliciously modified; that users may not understand the full authority delegated to an agent; and that post-incident investigators may need to reconstruct why an external action occurred.  We do not assume that the BOM by itself prevents attacks.  Instead, it gives reviewers a structured view of the authority envelope within which attacks and failures can become consequential.

In this model, an attacker may inject instructions into retrieved documents, poison tool metadata, exploit overbroad credentials, trigger an unsafe action through a legitimate tool, exploit missing approval gates, contaminate persistent memory, or abuse trust propagation in a multi-agent workflow.  These risks motivate the field groups in Table~\ref{tab:schema}.  For example, a RAG poisoning scenario needs source trust labels and retrieval logs; a tool-poisoning scenario needs tool source, descriptor hash, and review metadata; an excessive-agency scenario needs autonomy level, maximum tool tier, and approval rules.  The artifact is therefore a machine-readable representation of the security relationships among data, tools, authority, and evidence.

Table~\ref{tab:questions} gives the concrete review questions that \sys{} is designed to answer.  These questions are intentionally phrased in operational language because the intended users are security reviewers, platform engineers, governance teams, and incident responders.

\subsection{Authority-Envelope Schema}
Table~\ref{tab:schema} summarizes the major field groups. The schema is deliberately limited to fields that support review, comparison, or audit.  Fields can be marked unknown, not applicable, or inherited from external artifacts.  This is important for adoption: a BOM that requires complete disclosure of proprietary prompts or raw credentials would fail in practice.

\begin{table}[t]
\caption{Core \sys{} field groups.}
\label{tab:schema}
\centering
\scriptsize
\begin{tabular}{p{0.30\columnwidth}p{0.60\columnwidth}}
\toprule
\textbf{Layer} & \textbf{Security purpose} \\
\midrule
Agent identity & Owner, purpose, version, environment, business criticality. \\
Model layer & Provider, model, version, hosting mode, linked AI/ML-BOM. \\
Prompt-policy layer & Prompt hashes, change control, trusted boundaries, injection mitigations. \\
Tool layer & Tool source, protocol, descriptor, permissions, side effects, risk tier. \\
Memory-data layer & RAG sources, data classification, retention, vector store, retrieval logging. \\
Orchestration layer & Framework, max steps, code/browser/network access, sandboxing. \\
Autonomy-authority layer & Autonomy level, maximum tool tier, approval gates, emergency stop. \\
Inter-agent layer & Delegation, shared memory, trust domains, identity propagation. \\
Audit layer & Prompt, tool-call, retrieval, approval, and memory-write logs. \\
Provenance & Generated by, commit hash, timestamp, signature, external BOM references. \\
\bottomrule
\end{tabular}
\end{table}

\subsection{Risk Scoring Model}
The current implementation computes a transparent, rule-based score from six inputs: autonomy level, maximum tool-risk tier, data sensitivity, external exposure, memory persistence, and governance weakness.  Governance weakness increases when high-risk tools lack approval gates, logging is missing, emergency stop is absent, prompt changes are uncontrolled, or credentials are overbroad.  The score is not intended to replace expert review.  Its purpose is to sort agents, expose risk drivers, and make deployment drift visible.

\section{Implementation and Evaluation Corpus}
\subsection{Prototype Implementation}
The implementation contains a JSON Schema, YAML corpus files, a risk-scenario library, and a small command-line tool.  The core commands validate artifacts, compute scores, compare versions, map risk drivers to controls, and render reports.  The design deliberately avoids heavyweight infrastructure: no model training, no live exploit environment, and no dependency on proprietary enterprise systems.

\begin{figure}[t]
    \centering
    \includegraphics[width=\columnwidth]{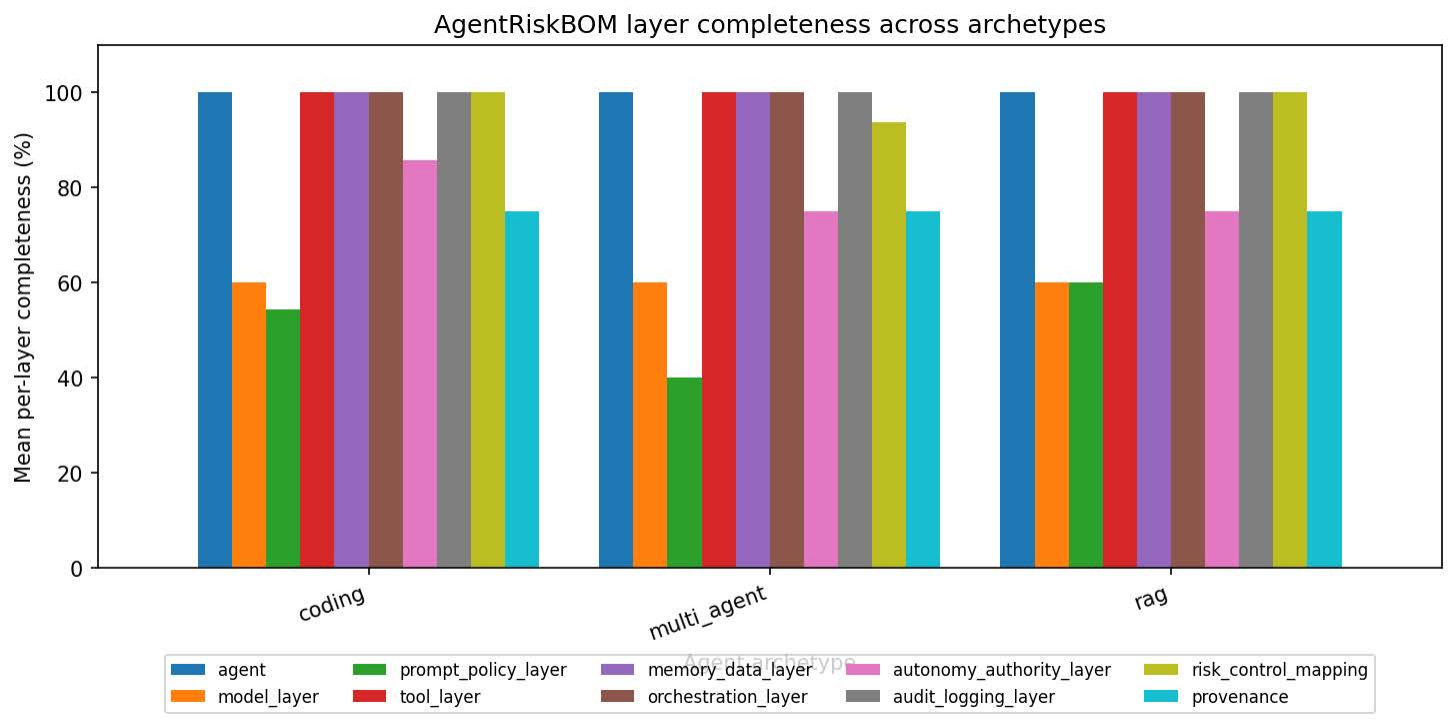}
    \caption{Schema expressiveness across corpus archetypes. Mean per-layer completeness exceeds 75\% for every archetype, while optional layers remain lower where they do not apply.}
    \label{fig:completeness}
\end{figure}

\subsection{Agent Corpus and Risk Scenarios}
The current corpus contains 13 real, documented open-source agents across three archetypes: coding agents, RAG agents, and multi-agent systems.  The agents are Aider, OpenHands, SWE-agent, Cline, Goose, Open Interpreter, AutoGPT, PrivateGPT, GPT-Researcher, MetaGPT, CrewAI, AutoGen, and BabyAGI.  The evaluation also uses 52 risk scenarios across 14 categories, including prompt injection, tool poisoning, excessive agency, sensitive-data disclosure, RAG poisoning, memory leakage, credential misuse, unsafe external action, inter-agent trust propagation, missing approval, missing audit logging, overprivileged cloud access, destructive tool misuse, and supply-chain compromise.

The risk-scenario library is not used as a synthetic attack benchmark; it is used as a coverage instrument.  Each scenario category asks whether a BOM view contains the minimum fields needed for a reviewer to notice that the risk applies.  For instance, \emph{unsafe external action} requires fields that describe tool side effects and approval gates, while \emph{memory leakage} requires fields for memory persistence, data classification, retention, and logging.  Table~\ref{tab:riskcats} lists the categories, required visibility fields and examples.

\begin{table*}[t]
\caption{Risk-scenario categories used to evaluate risk visibility.}
\label{tab:riskcats}
\centering
\scriptsize
\begin{tabular}{p{0.20\textwidth}p{0.31\textwidth}p{0.37\textwidth}}
\toprule
\textbf{Category} & \textbf{Example risk} & \textbf{Fields needed for visibility} \\
\midrule
Prompt injection & External content overrides task instructions. & Trusted input boundaries, prompt-injection mitigations, retrieval logs. \\
Tool poisoning & Tool metadata causes unsafe tool selection. & Tool source, descriptor hash, review date, allowed capability. \\
Excessive agency & Agent executes actions beyond intended scope. & Autonomy level, tool tier, approval requirements, emergency stop. \\
Sensitive-data disclosure & Agent exposes private or internal data. & Data classification, external communication, output controls, logs. \\
RAG poisoning & Retrieved documents alter agent behavior. & RAG source labels, provenance, retrieval logging, trust boundary. \\
Memory leakage & Persistent memory stores or reveals sensitive data. & Memory type, retention policy, memory-write logging, access control. \\
Credential misuse & Agent operates with overbroad credentials. & Credential scope, secret storage, rotation, least-privilege status. \\
Unsafe external action & Agent sends email, modifies systems, or triggers workflow. & Side effects, external endpoints, approval gates, audit logging. \\
Inter-agent trust & Delegated agents inherit unsafe authority. & Delegation policy, shared memory, trust domains, identity propagation. \\
Missing audit logging & Incident cannot be reconstructed. & Prompt, tool-call, retrieval, approval, and memory-write logs. \\
Missing approval & High-risk actions occur without human review. & Approval rules, tool-risk tier, autonomy level, approval logs. \\
Overprivileged cloud access & Agent uses cloud credentials broader than its task requires. & Credential scope, cloud permissions, least-privilege status, rotation policy. \\
Destructive tool misuse & Agent deletes, overwrites, executes, or modifies critical resources. & Destructive side effects, permission tier, sandboxing, approval gate, rollback evidence. \\
Supply-chain compromise & Agent, model, dependency, or tool artifact is replaced or tampered with. & External BOM references, artifact provenance, hashes, signatures, source registry, review metadata. \\
\bottomrule
\end{tabular}
\end{table*}

\section{Evaluation}
The evaluation asks six questions: what agentic dimensions are missing from prior BOMs; whether the schema is fillable across real agent archetypes; how much agentic risk becomes visible; whether risk-relevant deployment changes are detected; whether risk drivers can be mapped to controls and auditability; and whether the primary risk score is directionally consistent with a different scoring method. The study is artifact-centered: each experiment tests whether the schema and tooling make a specific class of security information visible or actionable.

\subsection{Prior-Art Coverage Gap}
Table~\ref{tab:coverage} compares 16 capability dimensions across SBOM, AI-BOM, ML-BOM, and \sys.  The matrix is derived from schema-level capability coverage rather than from how thoroughly any one artifact is populated.  \sys{} achieves a native-equivalent coverage total of 14 out of 16 capability dimensions.  SBOM, AI-BOM, and ML-BOM cover 1.0, 1.5, and 2.0 dimensions respectively under the same scoring.  This result is central because it identifies a structural gap: tool descriptors, tool permissions, autonomy, approval gates, credential scope, external action capability, and forensic readiness are not first-class fields in the prior BOM classes.

\begin{table}[t]
\caption{Prior-art capability coverage. Values: 0=no native coverage, 0.5=partial or indirect coverage, 1=native coverage.}
\label{tab:coverage}
\centering
\scriptsize
\setlength{\tabcolsep}{2.4pt}
\renewcommand{\arraystretch}{1.04}
\begin{tabular}{p{0.40\columnwidth}cccc}
\toprule
\textbf{Capability dimension} & \textbf{SBOM} & \textbf{AI-BOM} & \textbf{ML-BOM} & \textbf{\sys{}} \\
\midrule
Software dependency inventory & 1.0 & 0.0 & 0.0 & 0.5 \\
Model metadata & 0.0 & 1.0 & 1.0 & 0.5 \\
Dataset provenance & 0.0 & 0.5 & 1.0 & 0.0 \\
Prompt hierarchy / hashes & 0.0 & 0.0 & 0.0 & 1.0 \\
Tool descriptors and source & 0.0 & 0.0 & 0.0 & 1.0 \\
Tool permissions / risk tiers & 0.0 & 0.0 & 0.0 & 1.0 \\
Runtime autonomy level & 0.0 & 0.0 & 0.0 & 1.0 \\
Human approval gates & 0.0 & 0.0 & 0.0 & 1.0 \\
Memory persistence behavior & 0.0 & 0.0 & 0.0 & 1.0 \\
RAG source trust labels & 0.0 & 0.0 & 0.0 & 1.0 \\
Inter-agent communication & 0.0 & 0.0 & 0.0 & 1.0 \\
Credential scope & 0.0 & 0.0 & 0.0 & 1.0 \\
External action capability & 0.0 & 0.0 & 0.0 & 1.0 \\
Audit logging signals & 0.0 & 0.0 & 0.0 & 1.0 \\
Change diff with risk impact & 0.0 & 0.0 & 0.0 & 1.0 \\
Forensic readiness score & 0.0 & 0.0 & 0.0 & 1.0 \\
\midrule
\textbf{Native-equivalent total} & \textbf{1.0} & \textbf{1.5} & \textbf{2.0} & \textbf{14.0} \\
\bottomrule
\end{tabular}
\end{table}

\subsection{Schema Expressiveness Across Agent Archetypes}
All 13 corpus artifacts validate against the JSON Schema. Fig.~\ref{fig:completeness} shows that mean per-layer completeness exceeds 75\% for every archetype, while optional layers remain sparse where they do not apply, such as inter-agent mappings for RAG-only systems and vector-store retention metadata for coding assistants without long-term memory. This result supports the schema design goal: \sys{} can represent real agent archetypes without forcing non-applicable authority, memory, or delegation fields into every artifact.

\subsection{Agentic Risk Visibility}
\begin{figure}[t]
    \centering
    \includegraphics[width=\columnwidth]{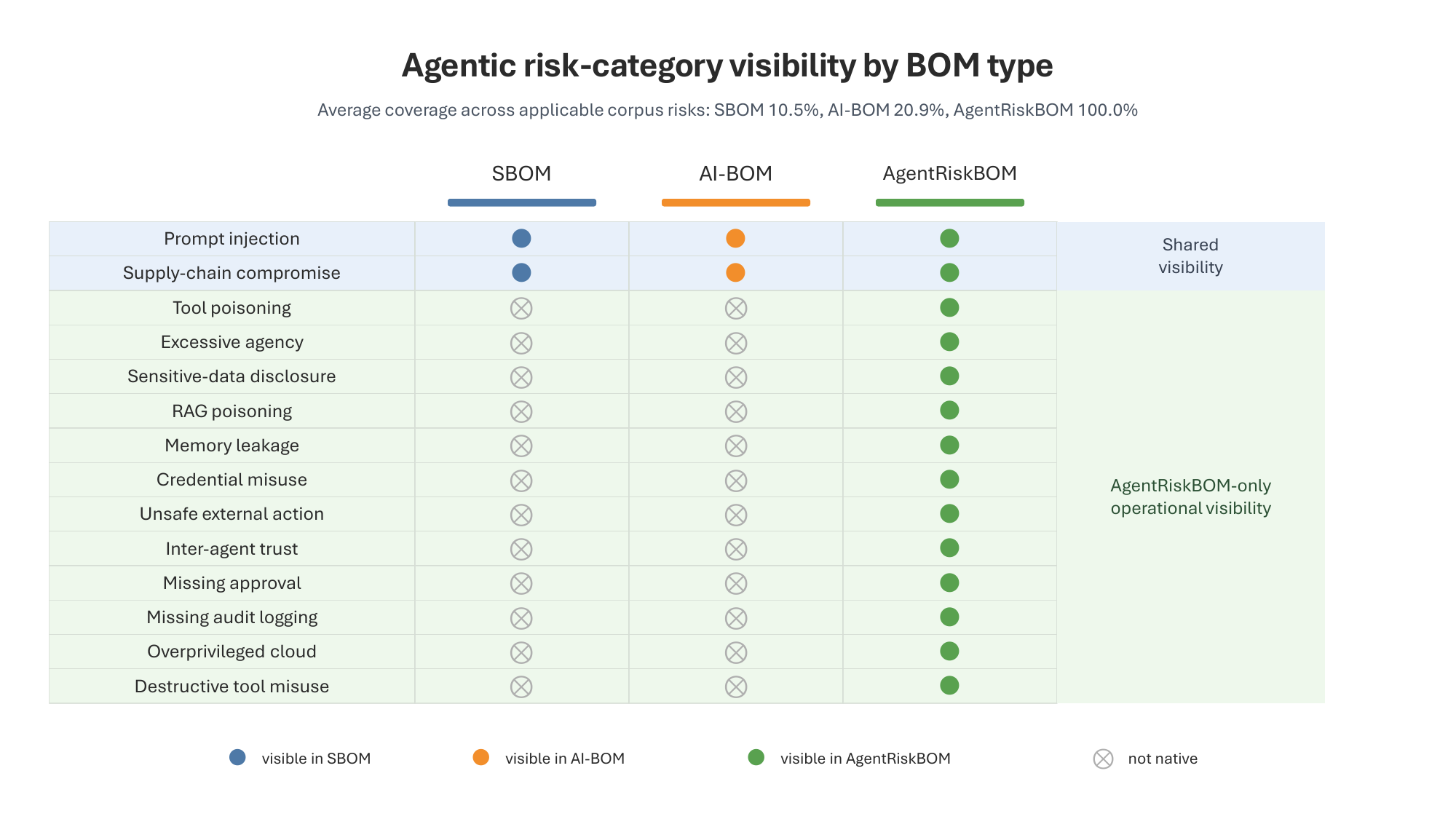}
    \caption{Agentic risk-category visibility across applicable corpus risks. This shows which specific categories are exposed by SBOM-like, AI-BOM-like, and \sys{} views.}
    \label{fig:riskcoverage}
\end{figure}

Fig.~\ref{fig:riskcoverage} compares average risk-category coverage.  Across applicable risks in the corpus, \sys{} exposes 100.0\% of modeled risk categories; SBOM-like and AI-BOM-like views expose 10.5\% and 20.9\%, respectively.  The categories visible only through \sys{} include excessive agency, unsafe external action, inter-agent trust, missing approval, missing audit logging, overprivileged cloud access, destructive tool misuse, credential misuse, and memory leakage.

\begin{figure*}[t]
    \centering
    \includegraphics[width=0.86\textwidth]{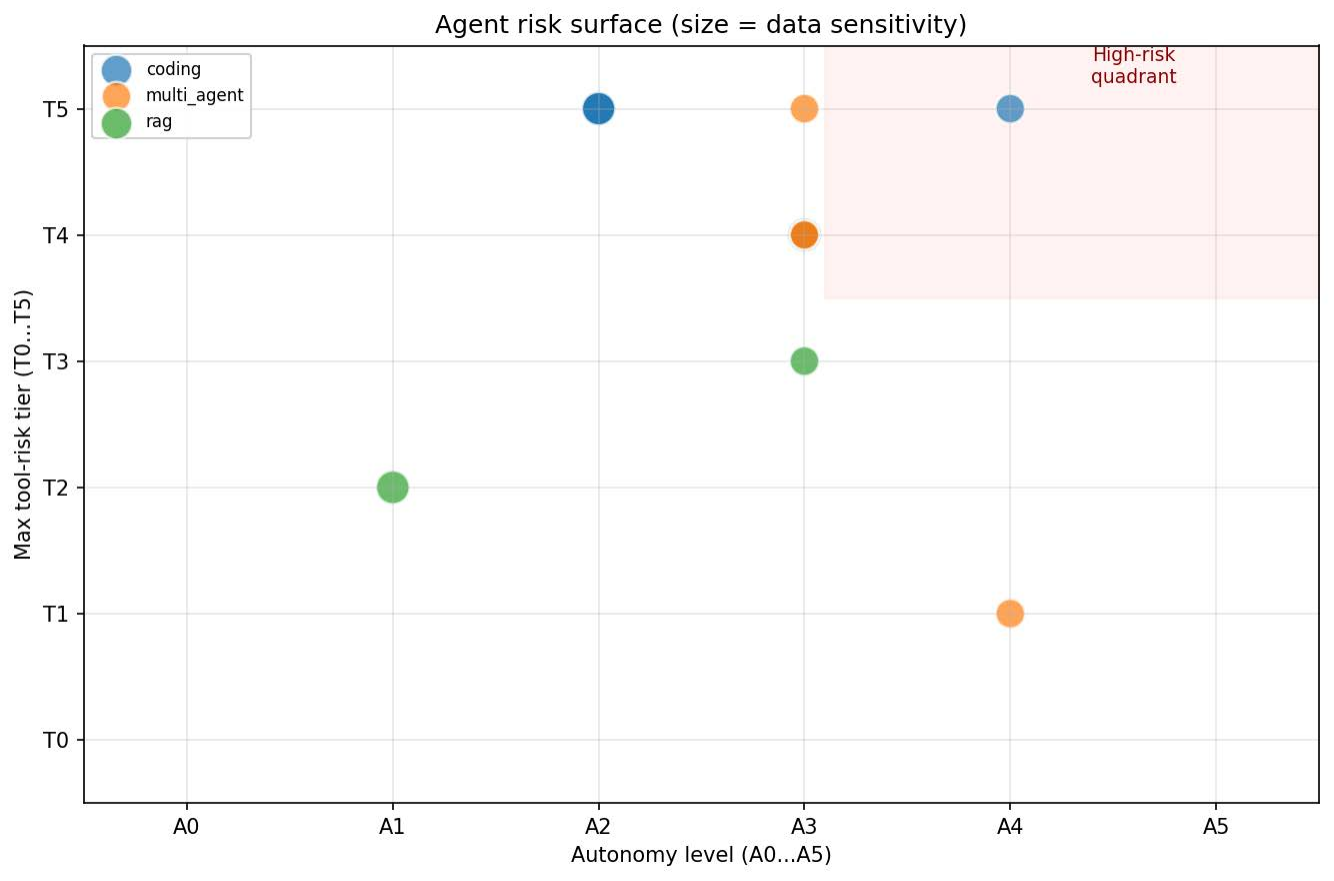}
    \caption{Agentic risk surface over autonomy level, maximum tool-risk tier, and data sensitivity. Seven of 13 corpus agents fall in the high-risk quadrant where autonomy is at least A3 and maximum tool-risk tier is at least T4.}
    \label{fig:surface}
\end{figure*}

Fig.~\ref{fig:surface} gives the main risk-surface result.  Seven of 13 agents fall into the high-risk quadrant where autonomy is at least A3 and maximum tool-risk tier is at least T4.  The high-risk quadrant is not a safety verdict; it is a review-priority signal for deployments where approval gates, least-privilege credentials, tamper-resistant logs, and change review matter most. In the risk-score distribution, the mean score is 53.5/100. One agent falls in the High band, none in Critical, and none in Low. AutoGPT is highest in the current run at 71.0, driven by autonomy A4, maximum tool tier T5, and internal data access.

\subsection{Agentic Authority Drift Detection}
Agentic authority drift is useful as a review concept only if risk-relevant changes can be detected before deployment. To test this property, each corpus artifact received up to three randomly selected structured mutations affecting declared authority, including tool additions, permission expansions, tier escalations, autonomy increases, memory-persistence increases, logging disablement, approval-gate removal, and model-provider changes. In total, 33 mutations were applicable and injected.

The diff detector identified all 33 mutations and assigned each to the correct change type. This result demonstrates complete recall for structured authority drift represented in the \sys{} schema, not general vulnerability discovery. That is the intended scope of a deployment gate over declared agent authority. The experiment shows that once tools, autonomy, memory, approvals, logging, and model metadata are represented as typed fields, risky configuration changes become mechanically reviewable across versions.

\subsection{Control Mapping and Forensic Readiness}
\begin{figure}[t]
    \centering
    \includegraphics[width=\columnwidth]{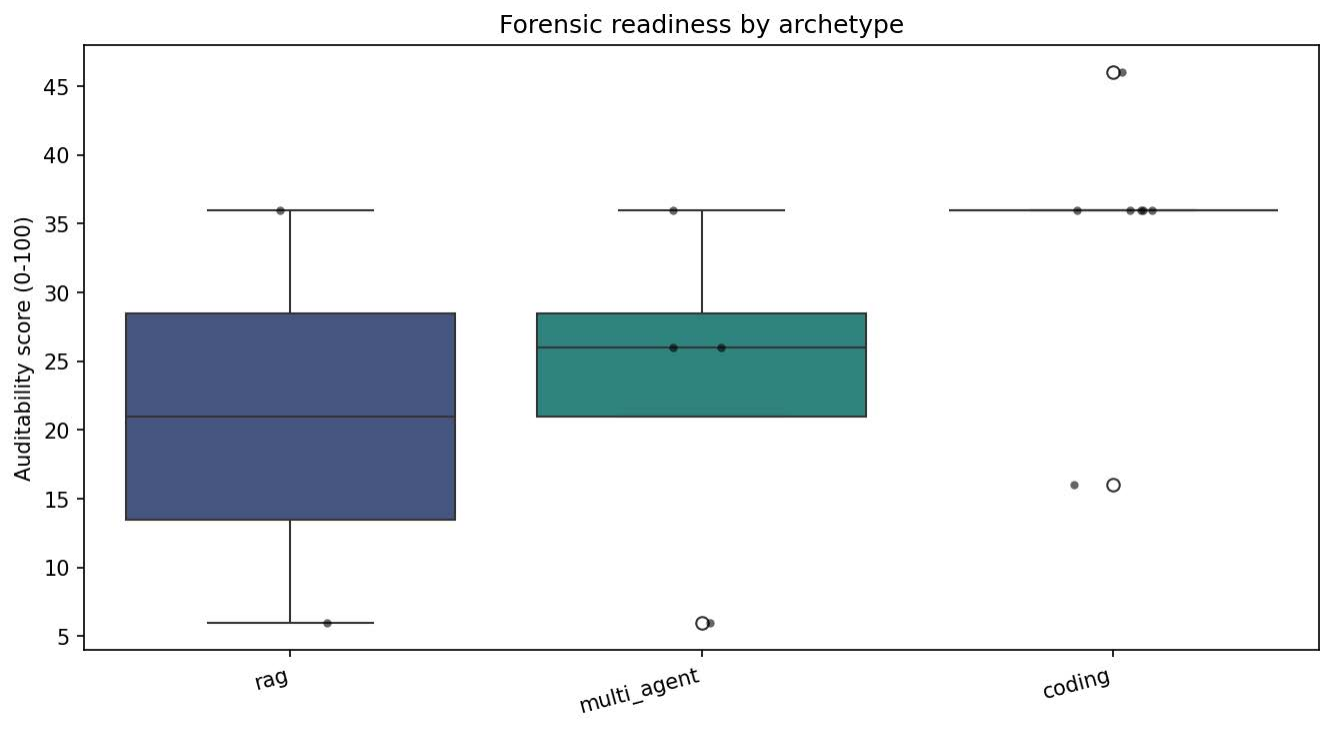}
    \caption{Forensic readiness by archetype. The auditability score combines logging signals, tamper resistance, log retention, prompt-change control, tool descriptor hashes, and provenance metadata.}
    \label{fig:auditability}
\end{figure}

The control mapper is rule-based: each rule reads named \sys{} fields and emits named control families.  High autonomy with T4--T5 tools maps to least privilege, human approval, emergency stop, and tamper-resistant logging; persistent memory over sensitive data maps to retention limits, retrieval logging, redaction, and access control; external communication maps to approval-before-send and domain allowlisting. Every risk-driver family surfaced by the scorer maps to at least one control family. Auditability remains weak across the corpus. Mean mapped control families per agent are highest for multi-agent systems (2.8) and lowest for coding agents (1.7). As shown in Fig.~\ref{fig:auditability}, coding agents have the highest mean auditability score (34.6/100), while RAG agents have the lowest mean score (21.0/100). This result indicates that many current agents expose authority without enough logging, retention, prompt-change control, descriptor hashing, or provenance metadata to reconstruct incidents reliably.

\subsection{Scoring Consistency and Calibration}
\begin{figure}[t]
    \centering
    \includegraphics[width=\columnwidth]{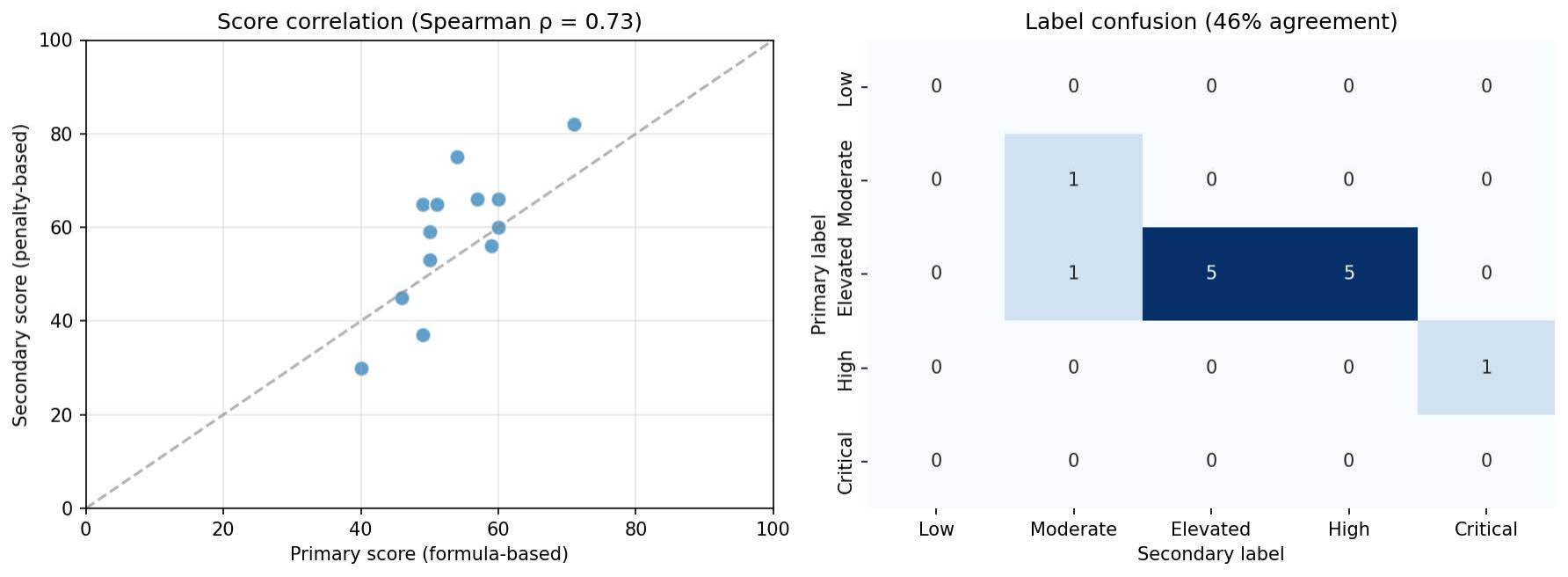}
    \caption{Primary score compared with an independent penalty-based scorer. The Spearman rank correlation is 0.73, while categorical label agreement is 46\%, indicating directional consistency but imperfect threshold calibration.}
    \label{fig:scorer}
\end{figure}

The secondary scorer is penalty-based and uses different weights from the primary formula.  Fig.~\ref{fig:scorer} shows that the Spearman rank correlation between primary and secondary scores is 0.73, indicating that the primary score is not merely an artifact of a single formula.  Categorical label agreement is 46\% across five labels, which should not be oversold.  It indicates that coarse labels such as Moderate, Elevated, and High require calibration, ideally with human security reviewers.  The score is therefore best used as a triage signal and change detector, not as an automated certification.

\section{Discussion}

\subsection{Delegated Authority as a Review Object} 
The evaluation establishes a structural result: a reviewable authority envelope can be represented, compared, and checked across real agent archetypes. Fields that are absent or only indirect in SBOM, AI-BOM, and ML-BOM views become explicit enough to support triage, drift review, control mapping, and auditability assessment. The high-risk quadrant in Fig.~\ref{fig:surface} illustrates this point as a scoping signal that identifies changes where review matters most. This is the practical contribution of \sys{}, which turns agent authority into a concrete object of review.

\subsection{Deployment Uses of \sys{}}
\sys{} is most useful at four points in the lifecycle. During procurement, it gives security teams a structured way to ask what an agent can access, remember, and change. During pre-deployment review, it exposes approval, credential, memory, and logging gaps before the agent is connected to production systems. During CI/CD, the diff tool can flag risky authority changes such as new destructive tools, increased autonomy, disabled logging, expanded credential scope, or persistent memory over sensitive data. During incident response, the auditability fields indicate whether enough evidence exists to reconstruct a harmful action.

\begin{figure*}[t]
    \centering
    \includegraphics[width=0.82\textwidth]{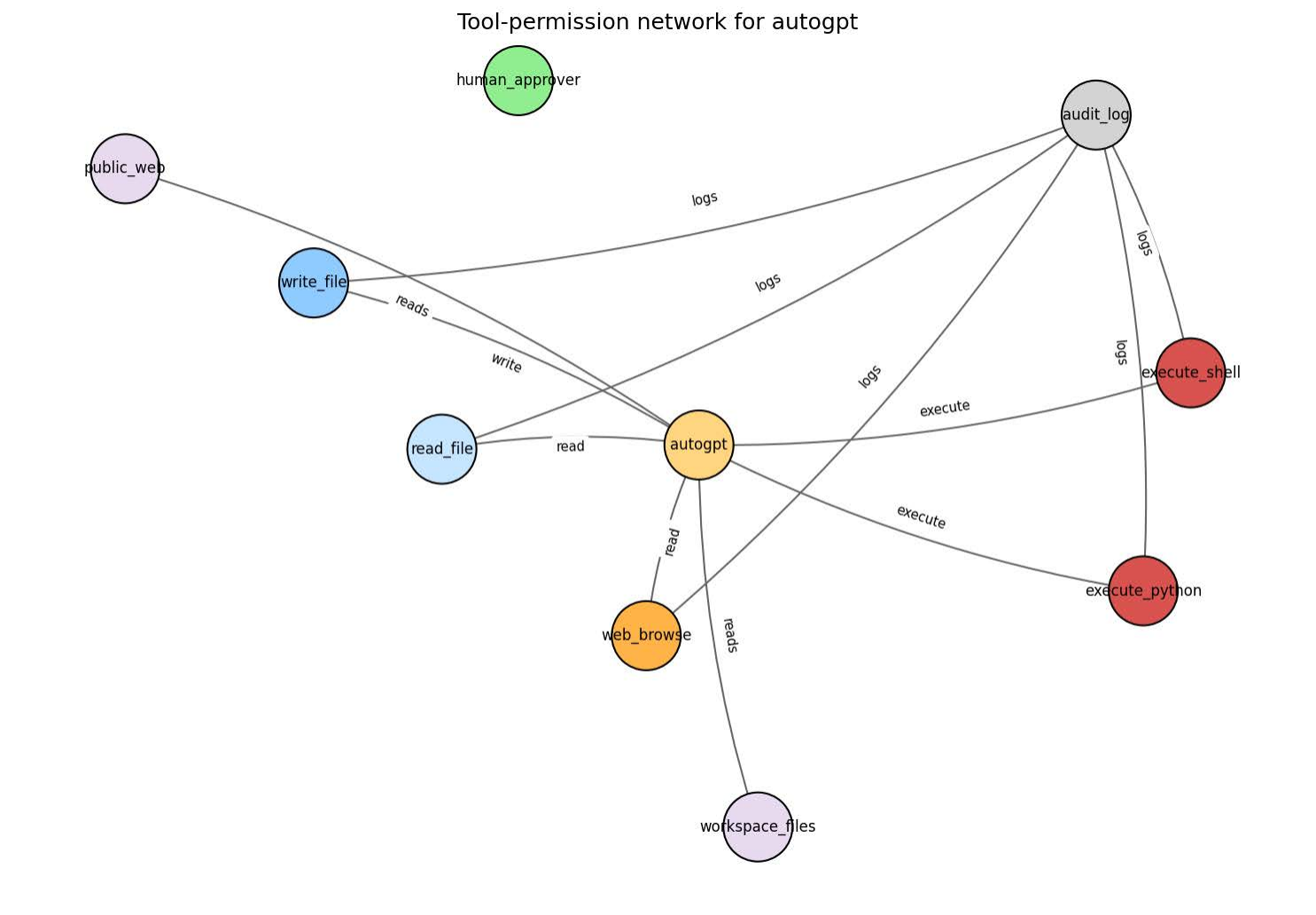}
    \caption{Tool-permission network for the highest-risk corpus agent, AutoGPT. The graph exposes the implicit trust structure among the agent, tools, external systems, audit log, and human approver.}
    \label{fig:network}
\end{figure*}

\subsection{Agentic Authority Drift}

Agentic authority drift is a deployment-level change in what an agent is able or permitted to do. Even when the base model, application code, and software dependency inventory appear unchanged, drift can occur through a new write-capable tool, broader credential scope, persistent memory, external communication, reduced approval requirements, modified tool metadata, or a changed orchestration policy. It is a distinct security problem for agentic systems because the risk-bearing object is not only the model or package set, but the authority assembled around the agent at runtime.

This drift is difficult to govern with conventional BOMs because authority is distributed across prompt templates, tool registries, API scopes, orchestration code, memory configuration, deployment policy, and human-in-the-loop controls. \sys{} makes this drift reviewable by representing the agent's authority envelope as a versioned artifact that reviewers can inspect, that CI/CD gates can compare before deployment, and that incident responders can use as a baseline for determining whether a deployed agent had the authority and evidence trail associated with a harmful action.

\subsection{Why a BOM Rather Than a Benchmark?}
Attack benchmarks and \sys{} answer different questions. A benchmark measures behavior under selected tasks or adversarial conditions. A BOM-style artifact describes the deployed configuration: tools, data reachability, approvals, credentials, memory, and audit evidence. Deployment teams need the latter when deciding whether a specific agent version should be connected to specific systems. For this reason, \sys{} is closer to a deployment-review object than to a leaderboard. It can be diffed, signed, archived, and attached to release decisions. It makes missing assumptions and risky authority changes visible before they become incidents. Fig.~\ref{fig:network} illustrates this deployment-level view for the highest-risk corpus agent.

\subsection{Standardization Path}
The simplest adoption path is to treat \sys{} as an agent-specific profile that references existing BOMs.  An SBOM remains the source of truth for packages, licenses, and vulnerability metadata; an AI-BOM remains the source of truth for model and training provenance; an \sys{} artifact records the delegated authority envelope over those components.  This profile-based path minimizes conflict with existing standards while leaving room for future extension namespaces for agent identity, tool descriptors, authorization scopes, memory stores, and audit evidence.

\subsection{Capability Opacity and Agentic AI Transparency}
We call the transparency gap exposed by agentic systems \textit{capability opacity}: the absence of a structured, reviewable account of what an agent can access, remember, change, delegate, and prove after the fact. Component inventories and model-provenance artifacts remain necessary, but they are insufficient when risk depends on runtime authority assembled from tools, data sources, credentials, memory stores, external systems, approval policies, and logs.

\sys{} supplies this missing layer by treating delegated authority as a first-class transparency object. It gives common names to tools, autonomy, memory, credentials, approval gates, inter-agent communication, audit signals, and external actions, while allowing lower-level software and model provenance to remain in SBOM, AI-BOM, and ML-BOM artifacts. The result is a practical capability-aware profile for reviewing, diffing, mapping, and auditing tool-using agents without requiring organizations to expose raw prompts, confidential documents, or proprietary execution traces.

\section{Limitations}
This study evaluates \sys{} as a risk-scoping and review artifact, not as a safety certificate for agents. The 13-agent corpus is sufficient to test schema expressiveness across coding, RAG, and multi-agent archetypes, while broader enterprise deployments may introduce additional fields, controls, and governance requirements. Likewise, the risk-scenario library is a coverage instrument: it measures whether a BOM view exposes the fields needed to recognize a class of risk, not whether a live exploit succeeds. The scoring and control-mapping components should be interpreted as triage aids. Their purpose is to surface risk drivers, prioritize human attention, and detect authority drift across versions; they do not replace environment-specific threat modeling, operational monitoring, or security judgment. The secondary scorer supports this interpretation: rank correlation is directionally useful, while categorical thresholds require calibration with practitioners and deployment evidence. Finally, \sys{} depends on accurate declarations about tools, credentials, memory, approval gates, and logging. Incomplete artifacts can obscure risk, as with any transparency mechanism. This boundary is also the reason the artifact is useful: by making delegated authority explicit, \sys{} gives reviewers a concrete object to validate, diff, archive, and govern before deployment, while complementary controls such as attestation, monitoring, and incident response provide assurance around it.

\section{Conclusion}
This paper introduced \sys{}, a lightweight security bill of materials for risk-scoping tool-using AI agents. The central claim is bounded: once an AI system can call tools, persist memory, access private data, coordinate with other agents, or affect external systems, its security posture cannot be inferred from software composition or model provenance alone. The delegated authority of the deployed agent must itself become a first-class object of review. 
The evaluation supports this claim by showing that \sys{} captures authority-related fields missing from prior BOM views, validates across a 13-agent open-source corpus, exposes modeled risk categories, detects structured authority drift, maps risk drivers to controls, and estimates forensic readiness. These results do not make \sys{} a safety certificate, and they do not replace runtime monitoring, incident response, or human security judgment. This work makes a practical contribution: \sys{} makes agent authority explicit, comparable, diffable, and governable before deployment.
As AI systems move from generating outputs to taking actions, security transparency must move from component inventory alone toward structured review of what agents are allowed to do.

\balance


\begin{thebibliography}{00}

\bibitem{b1} B. Xia, T. Bi, Z. Xing, Q. Lu, and L. Zhu, "An empirical study on software bill of materials: Where we stand and the road ahead." In 2023 IEEE/ACM 45th International Conference on Software Engineering (ICSE), pp. 2630-2642. IEEE, 2023.

\bibitem{b2} T. Stalnaker, N. Wintersgill, O. Chaparro, M. Di Penta, D. M. German, and D. Poshyvanyk, "Boms away! inside the minds of stakeholders: A comprehensive study of bills of materials for software systems." In Proceedings of the 46th IEEE/ACM International Conference on Software Engineering, pp. 1-13. 2024.

\bibitem{b3} M. Mitchell, S. Wu, A. Zaldivar, P. Barnes, L. Vasserman, B. Hutchinson, et al., "Model cards for model reporting." In Proceedings of the conference on fairness, accountability, and transparency, pp. 220-229. 2019.

\bibitem{b4} T. Gebru, J. Morgenstern, B. Vecchione, J. W. Vaughan, H. Wallach, H. Daumé III, et al., "Datasheets for datasets." Communications of the ACM 64, no. 12 (2021): 86-92.

\bibitem{b5} K. Bennet, G. K. Rajbahadur, A. Suriyawongkul, and K. Stewart, "Implementing ai bill of materials (ai bom) with spdx 3.0: A comprehensive guide to creating ai and dataset bill of materials." arXiv preprint arXiv:2504.16743 (2025).

\bibitem{b6} W. Vandendriessche, J. Thijsman, L. D'hooge, B. Volckaert, and M. Sebrechts, "AIBoMGen: Generating an AI Bill of Materials for Secure, Transparent, and Compliant Model Training." arXiv preprint arXiv:2601.05703 (2026).

\bibitem{b7} S. Yao, J. Zhao, D. Yu, N. Du, I. Shafran, K. Narasimhan, et al., ``ReAct: Synergizing Reasoning and Acting in Language Models,'' in \emph{Proc. International Conference on Learning Representations (ICLR)}, 2023.

\bibitem{b8} T. Schick, J. Dwivedi-Yu, R. Dessì, R. Raileanu, M. Lomeli, E. Hambro, et al., "Toolformer: Language models can teach themselves to use tools." Advances in neural information processing systems 36 (2023): 68539-68551.

\bibitem{b9} Q. Wu, G. Bansal, J. Zhang, Y. Wu, B. Li, E. Zhu, et al., "Autogen: Enabling next-gen LLM applications via multi-agent conversations." In First conference on language modeling. 2024.

\bibitem{b10} L. Wang, C. Ma, X. Feng, Z. Zhang, H. Yang, J. Zhang, et al., "A survey on large language model based autonomous agents." Frontiers of Computer Science 18, no. 6 (2024): 186345.

\bibitem{b11} X. Liu, H. Yu, H. Zhang, Y. Xu, X. Lei, H. Lai, et al., "Agentbench: Evaluating llms as agents." In International Conference on Learning Representations, vol. 2024, pp. 52989-53046. 2024.

\bibitem{b12} W. Jiang, N. Synovic, R. Sethi, A. Indarapu, M. Hyatt, T. R. Schorlemmer, et al., "An empirical study of artifacts and security risks in the pre-trained model supply chain." In Proceedings of the 2022 ACM workshop on software supply chain offensive research and ecosystem defenses, pp. 105-114. 2022.

\bibitem{b13} K. Greshake, S. Abdelnabi, S. Mishra, C. Endres, T. Holz, and M. Fritz, "Not what you've signed up for: Compromising real-world llm-integrated applications with indirect prompt injection." In Proceedings of the 16th ACM workshop on artificial intelligence and security, pp. 79-90. 2023.

\bibitem{b14} J. Yi, Y. Xie, B. Zhu, E. Kiciman, G. Sun, X. Xie, et al., "Benchmarking and defending against indirect prompt injection attacks on large language models." In Proceedings of the 31st ACM SIGKDD Conference on Knowledge Discovery and Data Mining V. 1, pp. 1809-1820. 2025.

\bibitem{b15} Q. Zhan, Z. Liang, Z. Ying, and D. Kang, "Injecagent: Benchmarking indirect prompt injections in tool-integrated large language model agents." In Findings of the Association for Computational Linguistics: ACL 2024, pp. 10471-10506. 2024.

\bibitem{b16} M. Mirakhorli, D. Garcia, S. Dillon, K. Laporte, M. Morrison, H. Lu et al., "A landscape study of open source and proprietary tools for software bill of materials (sbom)." arXiv preprint arXiv:2402.11151 (2024).

\bibitem{b17} S. Torres-Arias, H. Afzali, T. K. Kuppusamy, R. Curtmola, and J. Cappos, "in-toto: Providing farm-to-table guarantees for bits and bytes." In 28th USENIX Security Symposium (USENIX Security 19), pp. 1393-1410. 2019.

\bibitem{b18} S. Nathanson, A. Lee, C. C. Kieffer, J. Junkin, J. Ye, A. Saeed, et al., "AI Bill of Materials and Beyond: Systematizing Security Assurance through the AI Risk Scanning (AIRS) Framework." arXiv preprint arXiv:2511.12668 (2025).

\bibitem{b19} Y. Liu, G. Deng, Y. Li, K. Wang, Z. Wang, X. Wang, et al., "Prompt injection attack against llm-integrated applications." arXiv preprint arXiv:2306.05499 (2023).

\end{thebibliography}
\end{document}